**Blood Pressure Prediction for Coronary Artery Disease Diagnosis using Coronary Computed Tomography Angiography**

Rene Lisasi[1], Michele Esposito[2], Chen Zhao[1*]

1. Department of Computer Science, Kennesaw State University, Marietta, GA, 30060

2. Department of Cardiology, Medical University of South Carolina, Charleston, SC, USA

**Abstract**

Computational fluid dynamics (CFD)–based simulation of coronary blood flow provides valuable hemodynamic markers, such as pressure gradients, for diagnosing coronary artery disease (CAD). However, CFD is computationally expensive, time-consuming, and difficult to integrate into large-scale clinical workflows. These limitations restrict the availability of labeled hemodynamic data for training AI models and hinder broad adoption of non-invasive, physiology-based CAD assessment. To address these challenges, we develop an end-to-end pipeline that automates coronary geometry extraction from coronary computed tomography angiography (CCTA), streamlines simulation data generation, and enables efficient learning of coronary blood pressure distributions. The pipeline reduces the manual burden associated with traditional CFD workflows while producing consistent training data. We further introduce a diffusion-based regression model designed to predict coronary blood pressure directly from CCTA-derived features, bypassing the need for slow CFD computation during inference. Evaluated on a dataset of simulated coronary hemodynamics, the proposed model achieves state-of-the-art performance, with an $R^2$ of 64.42%, a root mean squared error of 0.0974, and a normalized RMSE of 0.154, outperforming several baseline approaches. This work provides a scalable and accessible framework for rapid, non-invasive blood pressure prediction to support CAD diagnosis.

## 1. Introduction

Coronary artery disease (CAD) is characterized by plaque build-up in the coronary arteries, which can restrict blood flow, impair cardiac function, and ultimately lead to heart failure [1]. As a result, CAD is the leading cause of death in the United States, accounting for approximately 610,000 deaths annually. In its early stages, CAD can often be managed with medication. However, when arterial narrowing becomes severe, interventions such as percutaneous coronary intervention (PCI) or coronary artery bypass grafting (CABG) are required to widen or bypass the affected vessels [2]. These procedures, while effective, carry additional risk for the patient.

Early identification of CAD can be facilitated by measuring the patient's fractional flow reserve (FFR) across a stenotic artery. An FFR ratio ≤ 0.80 indicates significantly reduced blood flow and compromised cardiac function, which may lead to heart failure [3]. FFR can be assessed invasively via coronary angiography (ICA) [4] or non-invasively using coronary computed tomography angiography (CCTA) [5]. ICA, while highly accurate, is invasive and involves guiding a thin catheter through the vasculature to the heart, posing considerable procedural risk [6]. In contrast, CCTA provides three-dimensional anatomical imaging of the coronary arteries with lower risk, requiring injection of a contrast agent before performing a CT scan [7].

To derive FFR from CCTA data, Fractional Flow Reserve Computed Tomography (FFRCT) is employed [7]. FFRCT models the coronary artery tree based on CT images and uses computational fluid dynamics (CFD) to simulate blood flow. Cardiologists can then visualize these simulations to assess the functional significance of stenoses and inform treatment planning, ranging from medication for low-risk cases to surgical intervention for high-risk patients.

While numerous studies have highlighted the diagnostic value of anatomical stenosis assessment via CCTA, misclassification remains a concern and may adversely affect treatment decisions. Consequently, there is a need for approaches that more accurately identify patients who would benefit from revascularization [3]. Performing CFD simulations from CCTA typically involves multiple steps: (a) loading CT images, (b) loading the 3D artery segmentation, (c) clipping the artery ends, (d) modeling the clipped artery, (e) meshing the model, (f) setting simulation parameters, (g) running the blood flow simulation, and (h) processing the simulation results, as illustrated in Figure 1. Detailed procedural descriptions are provided in Section A of the supplementary materials.

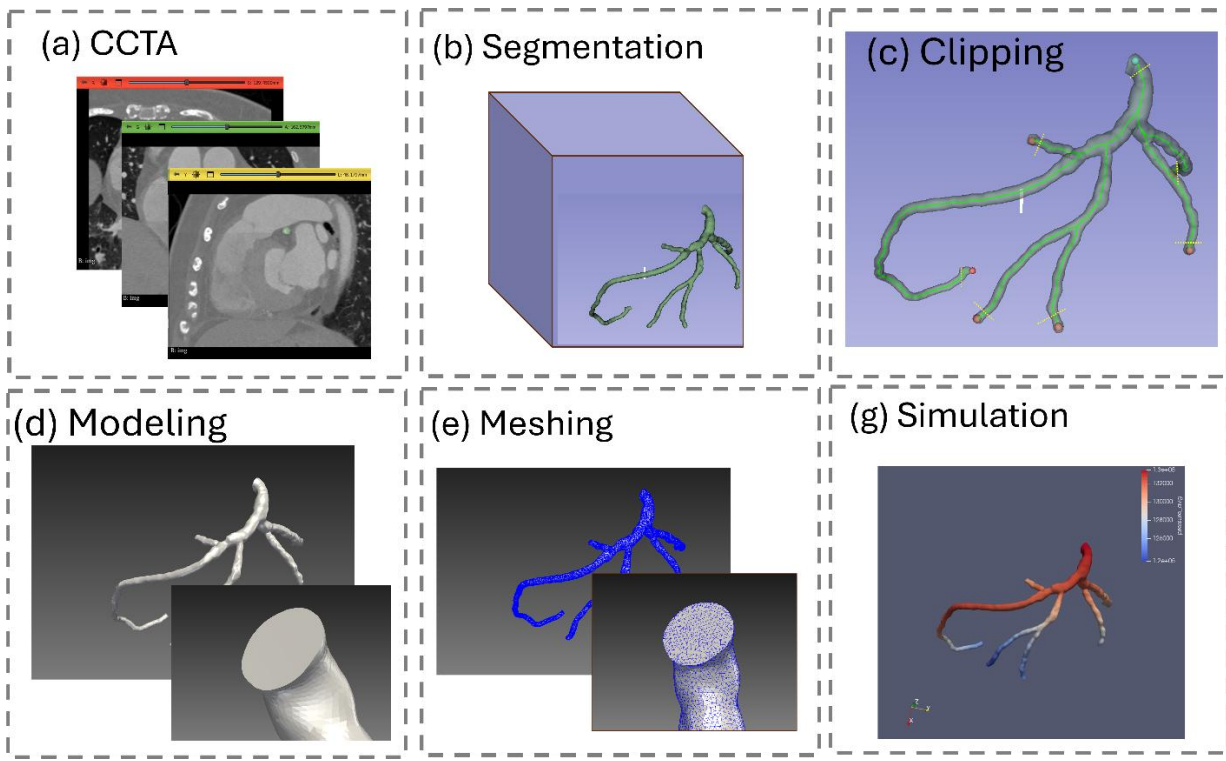

**Figure 1**. CCTA-based Blood Pressure Prediction. (a) CCTA scan, (b) 3D segmentation, (c) Clipping, (d) Modeling, (e) Meshing, (g) Blood flow simulation using SimVascular [8].

The proposed approach is two-fold, introducing both an innovative data pipelining method and a novel deep learning framework to support model training. Together, these methods establish a foundation for non-linear prediction in coronary artery medical imaging, specifically targeting blood pressure estimation. The pipelining framework streamlines dataset generation by automating artery segmentation and clipping, converting patient-specific coordinate systems to a common reference frame, and applying the required linear transformations. The ultimate goal is to predict blood pressure throughout the entire coronary artery tree for one or multiple patients, significantly reducing diagnostic costs and time, increasing clinical throughput and practitioner productivity, and minimizing patient risk.

## 2. Related Work

### 2.1 FFR evaluation using CCTA

Noninvasive calculation of FFR from CCTA ($FFR_{CT}$) is a novel method that applies CFD to determine the ischemia and the presence of CAD [9]. By employing mathematical models and numerical methods, CFD enables accurate prediction of blood flow patterns, velocities and pressures. These simulations provide valuable insights into the complex behavior of blood within arteries under various physiological and pathological conditions, offering a platform for investigating disease progression and evaluating potential therapeutic interventions. However, transient CFD analysis entails solving nonlinear partial differential equations with millions of degrees of freedom, typically demanding computational times ranging from 12 to 24 hours [10], which significantly limits the rapid hemodynamic calculation for emergency patients and real-time guidance in clinical practice. Moreover, HeartFlow [7,11,12] or ArteryFlow [13], while promising, are constrained by the CFD simulation framework, leading to limitations in both accuracy and computational speed. As a result, these methods have not been able to supplant the gold standard of invasive, $FFR_{wire}$, and cannot provide real-time guidance in clinical practice.

In contrast, we are proposing a diffusion-based method using deep learning to predict the blood flow pressure. With powerful computational resources, such as graphics processing units (GPUs), our FFR evaluation can be completed within 5 minutes. Additionally, the proposed deep learning technique serves as a low-risk and cost-effective alternative to invasive FFR. Consequently, it has the potential to greatly enhance clinical diagnostics and reduce risks associated with invasive interventions.

### 2.2 Diffusion Models for Medical Regression

A major challenge in coronary hemodynamics research is the lack of a publicly available, non-commercial toolkit that jointly provides CT images, detailed coronary artery trees, and corresponding blood-pressure fields derived from CFD simulations. This gap complicates model development and emphasizes the need for automated pipelines capable of generating training-ready datasets.

Traditional CFD-based FFR computation, although accurate, suffers from limited scalability. Each simulation requires segmentation, inlet and outlet clipping, mesh generation, parameter assignment, and multiple computationally intensive fluid-flow iterations. Performing this workflow manually for every patient is time-consuming, technically demanding, and costly. In contrast, deep learning–based regression offers scalability, minimal manual input, and near-instant inference once trained, making it an appealing alternative for population-scale CCTA screening.

Conventional regression architectures, including ResNet-based feature extractors and multi-layer perception (MLP) or long short-term memory regressors, serve as natural baselines for learning pressure along the centerline. ResNet's skip-connection structure maintains representational richness at greater depth [14], while MLPs provide a simple non-linear regressor (Obiora et al., 2023). LSTMs and Bi-LSTMs are commonly used for sequential prediction problems but introduce order-dependence across steps [15], which can propagate noise and degrade performance when the spatial ordering of centerline points does not align with temporal assumptions.

Diffusion models offer a fundamentally different paradigm. Originally developed for high-fidelity image synthesis (Ho et al., 2020), diffusion models consist of a forward process that gradually injects Gaussian noise into a sample and a reverse process that iteratively denoises a latent vector to recover a clean output. The reverse denoising process effectively learns a smooth, multi-step refinement trajectory, which is particularly advantageous for producing continuous-valued predictions. Although diffusion has been widely used for image generation, recent studies—such as GANs-Guided Conditional Diffusion Models for Synthesizing CT Images—demonstrate their capacity to handle conditional medical imaging tasks by leveraging the progressive denoising pathway.

Our approach adapts diffusion from generative modeling to continuous regression, reformulating the denoising trajectory as an iterative refinement process for predicting blood pressure. This yields smoother predictions than sequence-dependent methods (e.g., Bi-LSTM) and avoids the accumulation of ordering artifacts. Unlike Classification and Regression Diffusion (CARD) [16], which utilizes diffusion only to quantify uncertainty around predictions, our inverted conditional diffusion model directly performs regression and can be generalized to other continuous-valued medical prediction and classification tasks.

By modifying the diffusion framework to operate on regression targets rather than images, and by conditioning the model on CT-derived patch embeddings along the coronary centerline, our method provides a practical and scalable alternative to CFD for patient-specific hemodynamic estimation. This new class of diffusion-based regression models may help bridge the gap between purely image-based anatomical assessment and functional evaluation traditionally requiring invasive or computationally expensive procedures.

## 3. Methodology

The proposed method is made up of two main parts: the Patch-Based Dataset Pipeline (PBDP) and Inverted Conditional Diffusion (ICD) for blood flow prediction. The data pipeline consists of simulating blood flow in coronary arteries, extracting the centerline, image patches, and scalar pressures at each point for every ε into a dataset. The added improvement to the pipeline is converting the volume to Left Posterior Superior (LPS) coordinate system so that all files share a common coordinate system, then transforming the volume to the center of the pressure polygon so that it aligns the patches with the centerline, before finally converting world voxel coordinates to local coordinates.

### 3.1 Blood flow simulation using CCTA

The blood flow data for coronary arteries was generated through a multi-step pipeline involving data loading, artery clipping, centerline extraction, modeling, meshing, parameter assignment, simulation, and post-processing, as shown in Figure 1. Each step is detailed below.

1) Raw patient data, including CT scans and corresponding 3D segmentation labels, were imported into 3D Slicer. Within the Slicer interface, the segmentation editor was selected, and the 3D view enabled to visualize the coronary artery. The target artery was then clipped to isolate the region of interest.

2) Centerline Extraction. Using the VMTK module in Slicer, centerlines were extracted from the segmented arteries. The surface was set to the segmentation label, and new endpoints were defined. A new centerline model was generated and applied, with endpoint adjustments ensuring endpoints were only located at the termini of the artery tree.

3) Inlet and Outlet modification. To prepare the artery for CFD meshing, the inlet and outlet segments were clipped using the Clip Vessel module in 3D Slicer. The newly generated centerline and endpoints were used to define clipping locations. Caps were optionally generated for the inlet and outlet surfaces, but the final capping was deferred to the modeling step for precise boundary treatment.

4) Modeling. Using SimVascular, the clipped artery imported as a model. Face extraction was performed with a separation angle of 50°, and the global representation reinitialized. Surfaces not corresponding to anatomy were removed, holes were filled, and caps labeled according to their anatomical location (e.g., Left Anterior Descending, LAD).

5) Meshing. A new mesh was generated for the imported model. Mesh size was estimated, and the mesher executed. Successful meshing requires properly clipped inlets and outlets; if meshing fails, the clipping and capping procedures were adjusted. The model can then be re-meshed as necessary to ensure mesh quality.

6) Blood flow simulation. The following parameters were assigned to perform the blood flow simulation using SimVascular. Using the configured mesh and parameters, simulations were executed in SimVascular. Parallel computing with MPI was enabled to leverage available cores (10 cores in this setup). The simulation generated time-resolved hemodynamic data across the cardiac cycle. After performing the simulation, visualization and analysis were performed in ParaView, enabling playback of the entire cardiac cycle. Average pressures were extracted using the average mmhg field for downstream analysis.

- **Initial pressure:** 133,300 Pa (treated as 100 mmHg for results)
- **Boundary conditions:** All caps set to a resistance of 1333 Pa·s/cm³, except the inlet.
- **Wall properties:** Thickness = 0.2 mm, Elastic modulus = 4×10⁶ Pa, Density = 0.8 g/cm³, Pressure = 133,300 Pa
- **Solver parameters:** As summarized in Table 1.

**Table 1**. Solver parameters employed to simulate the blood flow using SimVascular

| Description | Value |
| --- | --- |
| Number of timesteps | 150 |
| Timestep size | 0.001 |
| Number of timesteps between restarts | 10 |
| Step construction | 4 |
| Residual Criteria | 0.001 |
| SvLS Type | GMRES |
| Tolerance on Momentum equations | 0.001 |
| Tolerance on Continuity Equations | 0.01 |
| Tolerance on NS Solver | 0.01 |
| Maximum Number of iterations for SvLS NS Solver | 10 |
| Maximum Number of iterations for SvLS Momentum Loop | 10 |

### 3.2. Patch-Based Data Preparation

The blood flow simulation generates pressure values on every face of the 3D artery mesh. However, fractional flow reserve evaluation does not require pressure information at every point on the vessel wall; it only depends on the pressure along the artery centerline, which mimics the clinical procedure of inserting a wire invasively to measure FFR at the center of the artery. Therefore, instead of using the full volumetric pressure data, we extract local 3D image patches around centerline points. These patches capture the imaging features of the surrounding vessel and are used as inputs to predict the centerline pressure, effectively linking anatomical imaging to hemodynamic function. This approach reduces data dimensionality, focuses on clinically relevant points, and aligns the deep learning model with real-world FFR measurement.

For each point on the centerline $C_m, m \in \mathbb{Z}^+$ of artery $n$, with volume origin $VO_n \in \mathbb{R}^3$ and voxel spacing $S_n \in \mathbb{R}^3$, we extract a 28×28×28 voxel patch $X \in \mathbb{R}^{D \times H \times W}$ centered at the point's coordinates. The average pressure within a radius $\epsilon = 5$ mm around the centerline point is computed and assigned to the patch. All coordinates are then converted to local coordinates relative to the artery centerline to ensure spatial consistency. Mathematically, the local voxel coordinates are calculated as shown in Eq. 1.

$$\begin{pmatrix} i_{n,m} \\ j_{n,m} \\ k_{n,m} \end{pmatrix} = \left( \begin{pmatrix} x_{n,m} \\ y_{n,m} \\ z_{n,m} \end{pmatrix} - VO_n * \begin{pmatrix} 1 \\ 1 \\ 1 \end{pmatrix} \right) * \frac{1}{S_n} \begin{pmatrix} 1 \\ 1 \\ 1 \end{pmatrix} \quad (1)$$

A prerequisite is that volumes, centerlines, and pressure data are all aligned in the same coordinate system to avoid mismatch, which is ensured by the volume alignment process. The volume alignment process is undertaken when a Right Anterior Superior (RAS) volume is converted to a Left Posterior Superior (LPS) compliant format. The process consists of converting the volume's RAS coordinates to LPS by flipping the sign of the x and y coordinates. Followed by translating the misaligned volume to the newly converted LPS coordinates' center. Then, rotating the translated volume by −180° if needed. The center is obtained by halving the product between the sum the origin and spacing, and the dimension.

For each artery, the algorithm iterates over the points along the centerline, extracting a 3D image patch centered at each point to capture local anatomical features. The corresponding pressure values are averaged within a small neighborhood around the centerline point, providing the target label for training. All coordinates are transformed into a local reference frame relative to the artery centerline, ensuring spatial consistency across different arteries and patients. This approach systematically constructs a dataset of paired imaging patches and centerline pressures, enabling the model to learn the mapping from anatomical features to hemodynamic function, effectively simulating FFR measurements from non-invasive imaging.

### 3.3. Inverted Conditional Diffusion

The proposed Inverted Conditional Diffusion (ICD) model is a modification of traditional conditional diffusion frameworks, repurposing the roles of input and label. Unlike standard diffusion models that generate images conditioned on labels, ICD treats the labels (here, pressure values) as input and the anatomical features (image patches) as conditioning, effectively inverting the diffusion process. This allows the network to regress blood pressure values from local imaging features in a manner that traditional encoder-only networks cannot replicate.

Forward diffusion process. In ICD, the forward process gradually adds Gaussian noise to the pressure labels over $T$ time steps, producing a noisy representation $y_T$. Formally, the forward process follows a Markov chain, as shown in Eqs. 2 and 3.

$$q(x_{1:T}|x_0) = \prod_{t=1}^{T} q(x_t|x_{t-1}) \qquad (2)$$

$$q(x_t|x_{t-1}) = \mathcal{N}(x_t; \sqrt{1-\beta_t} x_{t-1}, \beta_t I) \qquad (3)$$

where $x_0$ is the raw CT patch derived using the method in section 3.2. And $\beta_t \in (0,1)$ is a variance schedule that controls the amount of noise added at each step. After $T$ steps, $x_T$ becomes a nearly uniform Gaussian, i.e. $q(x_T|x_0) \approx \mathcal{N}(0,I)$.

**Reverse Diffusion Process**. The reverse process learns a parameterized distribution $p_\theta(y_{t-1} \mid y_t, x)$ conditioned on imaging features $x$, iteratively denoising $y_T$ back to the original pressure $y_0$. The reverse transitions are modeled as Gaussian distributions with learned means, as shown in Eq. 4.

$$p_\theta(x_{t-1}|x_t, y) = \mathcal{N}(x_{t-1}; \mu_\theta(x_t, t, y), \sigma_t^2 I) \qquad (4)$$

This denoising effectively predicts pressure values from the imaging features by reversing the forward noise process.

**Conditioning and Sampling.** Conditioning is achieved by concatenating the image patch features and relative centerline coordinates directly into the denoising network. At inference, the model starts from Gaussian noise and iteratively applies the learned reverse process to recover the predicted pressures at centerline points. This strategy allows ICD to construct a latent representation from the labels and work backward, providing accurate regression even with complex non-linear relationships.

The overall algorithm of ICD is defined as in Algorithm 1.

**ALGORITHM 1: Proposed Inverted Conditional Diffusion**

| | |
|---|---|
| 0 | **Input:** sample $y_0$, conditioning variable $c$, noise schedule $\{\beta_t\}_{t=1}^T$ |
| 1 | Define $\alpha = 1 - \beta_t$ and $\bar{\alpha}_t = \prod_{s=1}^t \alpha_s$ |
| 2 | **//Forward Process** |
| 3 | For t=1 to T do: |
| 4 |    Sample $\epsilon \sim \mathcal{N}(0, I)$ |
| 5 |    $y_t = \sqrt{\bar{\alpha}_t} y_0 + \sqrt{1 - \bar{\alpha}_t} \epsilon$ |
| 6 | End for |
| 7 | **//Reverse Process** |
| 8 | Train conditional network $\epsilon_\theta(y_t, t, c)$ to predict $\epsilon$ |
| 9 | For t=T to 1 do: |
| 10 |    Sample $z \sim \mathcal{N}(0, I)$ if $t > 1$, else $z = 0$ |
| 11 |    Compute: |
| | $$y_{t-1} = \frac{1}{\sqrt{\alpha_t}} \left( y_t - \frac{\beta_t}{\sqrt{1 - \bar{\alpha}_t}} \epsilon_\theta(y_t, t, c) \right) + \sigma_t z$$ |
| 12 | End for |
| | Ensure: Denoised sample $y_0$ |

**Architecture.** The architecture of the proposed ICD is shown in Figure 2. The proposed Inverted Conditional Diffusion (ICD) model is designed to predict blood pressure at each coronary artery centerline point by integrating local anatomical information with geometric positional cues. To achieve this, the framework begins with a 3D convolutional encoder that processes a volumetric CCTA patch centered around the vessel lumen. This encoder consists of sequential 3D convolution, batch normalization, ReLU activation, and max-pooling layers, progressively extracting multi-scale spatial features. After the final convolutional block, the resulting feature map is flattened and projected into a compact 128-dimensional embedding that captures local vessel morphology, lumen intensity patterns, plaque burden, and surrounding tissue characteristics.

In parallel, each centerline coordinate $(x, y, z)$ is encoded through a lightweight linear mapping to produce a 3-dimensional geometric embedding. This embedding provides the model with structural context, enabling it to learn how blood pressure naturally varies along the artery according to vessel curvature, branch location, and proximity to stenotic regions. The anatomical and geometric embeddings are concatenated to form a unified 288-dimensional representation describing the imaging context and spatial identity of each target point. This fused representation serves as conditioning input for the diffusion process.

The diffusion module forms the core of the ICD model. Unlike conventional diffusion models used for image generation, our formulation inverts the process to operate directly in the regression domain. During training, Gaussian noise is progressively added to the ground-truth pressure values, forming a forward diffusion trajectory. The model then learns the reverse denoising trajectory, where a neural network—conditioned on the fused anatomical–geometric representation—iteratively predicts a cleaner pressure estimate at each diffusion timestep. Two fully connected layers with ReLU activation serve as the denoising backbone, and a final linear layer outputs a single pressure value representing the FFR-equivalent prediction at that centerline point. The ICD design eliminates sequential dependencies found in recurrent models and yields smooth, physically consistent pressure predictions that align more closely with coronary physiology.

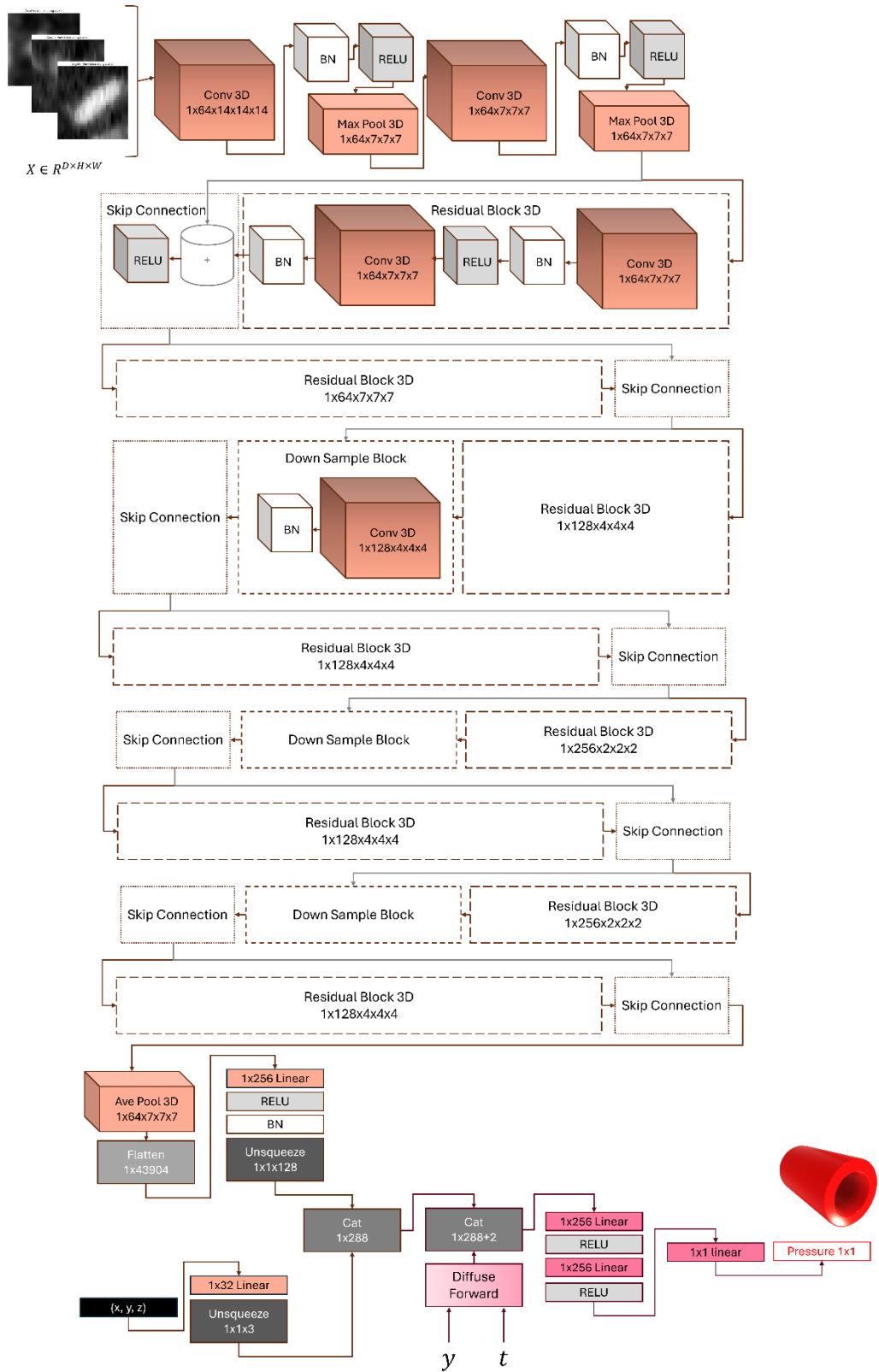

**Figure 2**. Architecture of the proposed ICD for blood flow prediction using patched CCTA images.

### 3.4. Loss Function and Optimization

We employed the Huber Loss to optimize the proposed model. Huber loss function outperforms other loss functions in time-series analysis by combining MSE and MAE, choosing MSE for small errors and MAE for large errors, providing more stable results [17].

The Huber loss function is as follows where $\delta$ is a threshold parameter, as shown in Eq. 5.

$$L_\delta(y, f(x)) = \begin{cases} \frac{1}{2}(y - f(x))^2, & for |y - f(x)| \leq \delta \\ \delta|y - f(x)| - \frac{1}{2}\delta^2, & otherwise \end{cases} \tag{5}$$

The optimizer used is Adam with decoupled weight decay for its ability to render optimal settings of the learning rate and decoupling weight decay, removing unnecessary hyperparameter tuning, speeding up convergence, and reducing overfitting.

### 3.5 Evaluation Metrics

The evaluation metrics compiled in this study are comprised of R-squared, Pearson correlation, normalized root mean squared error, root mean squared error. R-squared is the coefficient of determination and is the proportion of dependent variable variance, predictable from independent variables with the equation:

$$R^2 = 1 - \frac{\sum_{i=1}^{m}(\hat{y}_i - y_i)^2}{\sum_{i=1}^{m}(\bar{y} - y_i)^2} \tag{6}$$

where $y_i$ indicates the ground truth value of the blood pressure for one patch in our dataset and $\hat{y}_i$ is the model prediction. $\bar{y}$ indicates the average pressure values among the extracted patches.

Pearson Correlation coefficient provides a measure of the strength of linear relationship between two continuous random variables and is adopted when the data follows a normal data distribution, as defined in Eq. 7.

$$PCC = \frac{\sum((\hat{y}_i - \bar{\hat{y}}_i)^2)((y_i - \bar{y})^2)}{\sqrt{\sum(\hat{y}_i - \bar{\hat{y}}_i)^2}\sqrt{\sum(y_i - \bar{y})^2}} \tag{7}$$

Root mean squared error (RMSE) is the square root of mean squared error, a common machine learning evaluation metric. Taking the square root of MSE keeps the metric close to values already being worked with, making it more relative than the ballooning MSE with the following equation:

$$RMSE = \sqrt{\frac{1}{m}\sum_{i=1}^{m}(\hat{y}_i - y_i)^2} \tag{8}$$

Normalized root mean squared error (NRMSE) is the normalized RMSE, facilitating the comparison between models with differing scales by connecting RMSE to the observed range (Jadon et al., 2022) with the following equation:

$$NRMSE = \frac{\sqrt{\frac{1}{m}\sum_{i=1}^{m}(\hat{y}_i - y_i)^2}}{\bar{o}} \tag{9}$$

Here, $m$ is the number of data samples $\bar{o}$ is the average observation of the value.

## 4. Experimental Results

### 4.1 Enrolled Datasets

We employed two datasets in this study: CCTA1000 and CCTA36. CCTA1000 corresponds to the publicly available ImageCAS dataset [18], which contains high-quality coronary CT angiography (CCTA) scans with expertly annotated coronary artery segmentations. CCTA36 is a private dataset derived from SPECT MPI studies [19] and includes patients with invasive fractional flow reserve (FFR) measurements.

The CCTA1000 dataset consists of 3D CCTA scans acquired using a Siemens 128-slice dual-source CT scanner from one thousand patients. For individuals previously diagnosed with coronary artery disease, early revascularization (within 90 days) was included. During reconstruction, either the 30%–40% or 60%–70% cardiac phase was selected to ensure optimal visualization of coronary arteries. The resulting scans have a spatial resolution of 512 × 512 × (206–275) voxels.

The CCTA36 dataset includes thirty-six patients with at least one coronary stenosis ⩾50%. These cases were retrospectively collected, and each patient underwent SPECT MPI imaging followed by invasive FFR assessment, providing ground-truth functional measurements.

For experiments, we used a subset of each dataset. From CCTA1000, 40 scans were used for training, 5 for validation, and 10 for testing. From CCTA36, 10 of of 36 patients were enrolled, and 6 scans were used for training, 2 for validation, and 2 for testing. CCTA1000 is considered a medium-sized dataset, whereas CCTA36 represents a small, functionally annotated dataset.

### 4.2 Implementation details

The CT-scans and 3D segmentations were used to simulate the data, generating pressure files, after which centerlines files are extracted. The set of centerlines, pressures, and CT-scans are used to generate the train and test dataset. PyTorch is used to implement deep regression models. The hardware specifications are NVIDIA RTX 4090, 64GB RAM, and Intel i9. The software used for deep learning were Python 3.11.9, Py-Torch 2.4.1, while the simulation software included 3D Slicer 5.6.2, Paraview 5.13.0 (Ahrens et al, 2005), Fortran 2024.2.1.1084, MPI and Simvascular 2023-03-27. The Vedo library was used to visualize the transformation changes in an unbiased plane. The K3D library was used to visualize pressure predictions.

### 4.3 Results of blood flow prediction

To assess the performance of the proposed Resnet50-ICD model, it is compared to CNN-MLP, CNN-Attention-BiLSTM, CNN-ICD, Resnet50-MLP, Resnet50-Attention-BiLSTM all implemented in PyTorch with a random seed of 42. The following are the architectures of the baseline models.

- **CNN–MLP (CM):** The CNN–MLP baseline predicts blood pressure at each centerline point by jointly leveraging local image features and spatial geometry. A 3D CNN encoder extracts features from a cubic patch centered on the coronary artery using three convolutional blocks with batch normalization, ReLU activations, and max-pooling, progressively capturing morphological details of the lumen and surrounding tissue. The resulting feature map is flattened and projected into a 128-dimensional representation through a fully connected layer with normalization and dropout. Simultaneously, the (x, y, z) coordinates of the centerline point are processed by a lightweight MLP, mapping the geometric position into a 32-dimensional embedding. The two feature vectors are concatenated and passed through a two-layer regression head to produce the final scalar pressure estimate. This hybrid architecture effectively fuses anatomical appearance with spatial location, providing a strong deterministic baseline for comparison with the proposed diffusion-based model.

- **CNN–Att-BiLSTM (CL):** The CNN–Attention Bi-LSTM (CNN–Att) model extracts 3D image features using a CNN, integrates them with centerline coordinates, and employs a Bi-LSTM with attention to capture sequential dependencies along the vessel. The attended features are then fed into a regression head to predict blood pressure at each centerline point.

- **CNN–ICD (CD):** In this variant, the ResNet in the proposed model is replaced with plain CNN layers.

- **ResNet50–MLP (RM):** This model replaces the diffusion module with a standard MLP.

- **ResNet50–Att (RL):** Here, the CNN layers in the CNN–Att model are replaced with ResNet-50 to extract deeper representations from the patched CCTA images.

Using the above baselines and the same data splitting strategy, we validated the proposed model using CCTA1000 dataset. The results are shown in Table 2 and Table 3.

**Table 2**. R-squared & Pearson Correlation of case-wise evaluation using subjects in CCTA1000

| Test Case Number | RD (ours) | | RL | | RM | | CD (ours) | | CL | | CM | |
|---|---|---|---|---|---|---|---|---|---|---|---|---|
| | $R^2$ | Pearson | $R^2$ | Pearson | $R^2$ | Pearson | $R^2$ | Pearson | $R^2$ | Pearson | $R^2$ | Pearson |
| 1 | 66.04 | 85.34 | 55.33 | 91.57 | 65.33 | 84.83 | 52.65 | 81.36 | 44.54 | 87.49 | 42.3 | 79.49 |
| 2 | 71.41 | 87.52 | 68.45 | 85.12 | 76.05 | 89.25 | 82.21 | 91.7 | 54.2 | 89.73 | 72.06 | 87.15 |
| 3 | 54.52 | 86.11 | 22.22 | 82.56 | 66.91 | 87.87 | 71.46 | 90.1 | 9.34 | 85.2 | 61.36 | 87.48 |
| 4 | 71.97 | 85.84 | 26.25 | 78.28 | 72.41 | 85.17 | 76.8 | 87.66 | 75.13 | 88.04 | 75.51 | 87.18 |
| 5 | 79.67 | 89.49 | 34.81 | 83.1 | 56.06 | 79.26 | 81.06 | 90.1 | 76.6 | 89.9 | 61.01 | 80.36 |
| 6 | 74.48 | 89.84 | 49.45 | 83.24 | 65.84 | 86.07 | 69.39 | 90.16 | 70.68 | 86.11 | 72.99 | 87.77 |
| 7 | 39.66 | 76.19 | 9.09 | 74.25 | 26.04 | 71.77 | -7.27 | 60.17 | 50.83 | 75.18 | -10.6 | 58.96 |
| 8 | 58.28 | 78.22 | 69.06 | 89.51 | 19.42 | 78.48 | 48.19 | 75.11 | 26.89 | 86.16 | 68.6 | 84.48 |
| 9 | 68.12 | 89.76 | 78.01 | 89.97 | 76.21 | 92 | 69.75 | 91.96 | 79.79 | 90.08 | 77.95 | 91.37 |
| 10 | 60 | 81.47 | 33.76 | 83.65 | 65.94 | 85.8 | 66.69 | 88.24 | 77.83 | 89.99 | 70.87 | 88.73 |
| **Mean** | **64.42** | 84.978 | 44.64 | 84.125 | 59.021 | 84.05 | 61.093 | 84.656 | 56.583 | **86.788** | 59.202 | 83.297 |

**Table 3**. RMSE & NRMSE of case-wise evaluation using subjects in CCTA1000

| Test Case Number | RD (ours) | | RL | | RM | | CD (ours) | | CL | | CM | |
|---|---|---|---|---|---|---|---|---|---|---|---|---|
| | RMSE | NRMSE | RMSE | NRMSE | RMSE | NRMSE | RMSE | NRMSE | RMSE | NRMSE | RMSE | NRMSE |
| 1 | 0.81 | 0.1351 | 0.93 | 0.1549 | 0.82 | 0.1365 | 0.96 | 0.1595 | 1.03 | 0.1726 | 1.05 | 0.1726 |
| 2 | 0.86 | 0.1265 | 0.91 | 0.1329 | 0.79 | 0.1158 | 0.68 | 0.0998 | 1.09 | 0.1601 | 0.85 | 0.125 |
| 3 | 0.81 | 0.1795 | 1.06 | 0.2348 | 0.69 | 0.1531 | 0.64 | 0.1422 | 1.15 | 0.2534 | 0.75 | 1.15 |
| 4 | 1.06 | 0.1446 | 1.72 | 0.2346 | 1.05 | 0.1435 | 0.96 | 0.1316 | 1.00 | 0.1362 | 0.99 | 0.1352 |
| 5 | 0.8 | 0.1195 | 1.43 | 0.214 | 1.18 | 0.1757 | 0.77 | 0.1154 | 0.86 | 0.1282 | 1.11 | 0.1655 |
| 6 | 0.79 | 0.1387 | 1.11 | 0.1951 | 0.91 | 0.1604 | 0.86 | 0.1519 | 0.85 | 0.1486 | 0.81 | 0.1426 |
| 7 | 1.21 | 0.2118 | 1.48 | 0.26 | 1.33 | 0.2345 | 1.61 | 0.2824 | 1.09 | 0.1912 | 1.63 | 0.2868 |
| 8 | 0.89 | 0.1665 | 0.77 | 0.1434 | 1.24 | 0.2314 | 0.99 | 0.1856 | 1.18 | 0.2204 | 0.77 | 0.1445 |
| 9 | 1.23 | 0.1454 | 1.03 | 0.1207 | 1.07 | 0.1256 | 1.2 | 0.1416 | 0.98 | 0.1157 | 1.03 | 0.1209 |
| 10 | 1.28 | 0.1726 | 1.65 | 0.2222 | 1.18 | 0.1593 | 1.17 | 0.1575 | 0.95 | 0.1285 | 1.09 | 0.1473 |
| **Mean** | **0.974** | **0.154** | 1.21 | 0.19126 | 1.026 | 0.16358 | 0.984 | 0.15675 | 1.018 | 0.1655 | 1.008 | 0.25904 |

In Table 4 and 5, R-squared, Pearson correlation, RMSE, and NRMSE of CCTA36 cases are compared. It is apparent that the FFR dataset is extremely small and inconclusive but does give some insight into the performance given a limited dataset. Here, the Resnet models each win at least one category. An insight gathered is that proposed diffusion models are data intensive.

**Table 4**. R-squared & Pearson Correlation of case-wise evaluation using subjects in CCTA36

| Test Case Number | RD (ours) | | RL | | RM | | CD (ours) | | CL | | CM | |
|---|---|---|---|---|---|---|---|---|---|---|---|---|
| | $R^2$ | Pearson | $R^2$ | Pearson | $R^2$ | Pearson | $R^2$ | Pearson | $R^2$ | Pearson | $R^2$ | Pearson |
| 1 | 14.76 | 60.43 | 45.35 | 75.91 | 42.88 | 70.35 | -25.5 | 53.16 | 22.59 | 73.92 | 11.64 | 74.61 |
| 2 | 31.85 | 81.49 | 21.49 | 49.61 | 22.93 | 53.86 | 21.18 | 54.7 | 12.77 | 46.36 | 32.09 | 59.03 |
| Mean | 23.305 | 70.96 | 33.42 | 62.76 | 32.905 | 62.105 | -2.15 | 53.93 | 17.68 | 60.14 | 21.865 | 66.82 |

**Table 5**. RMSE & NRMSE of case-wise evaluation using subjects in CCTA36

| Test Case Number | RD (ours) | | RL | | RM | | CD (ours) | | CL | | CM | |
|---|---|---|---|---|---|---|---|---|---|---|---|---|
| | RMSE | NRMSE | RMSE | NRMSE | RMSE | NRMSE | RMSE | NRMSE | RMSE | NRMSE | RMSE | NRMSE |
| 2 | 3.65 | 0.2413 | 2.92 | 0.1932 | 2.99 | 0.1975 | 4.43 | 0.2928 | 3.48 | 0.23 | 3.72 | 0.2457 |
| 13 | 8.06 | 0.2008 | 8.65 | 0.2155 | 8.57 | 0.2135 | 8.67 | 0.2159 | 9.12 | 0.2272 | 8.04 | 0.2004 |
| Mean | 5.855 | 0.22105 | 5.785 | **0.20435** | **5.78** | 0.2055 | 6.55 | 0.25435 | 6.3 | 0.2286 | 5.88 | 0.22305 |

The results illustrate that the ResNet Inverse Conditional Diffusion (ICD) model achieved the highest average R-squared of 64.42%, RMSE of 0.974, and NRMSE of 0.154. The model also achieved the second-highest Pearson correlation of 84.978 on the medium-sized CCTA1000 dataset, along with the smoothest visual results closely resembling the ground truth. The CNN Inverse Conditional Diffusion model achieved the second-highest performance, with an average R-squared of 61.093, RMSE of 0.984, and NRMSE of 0.157 on the same dataset. It ranked third in terms of Pearson correlation at 84.656. These results demonstrate that the proposed Inverse Conditional Diffusion model achieves state-of-the-art performance in blood pressure prediction.

This performance suggests that the proposed model may also be effective for other non-linear regression and classification tasks. The best model for a small dataset was inconclusive, as different models excelled in isolated metrics. Due to order dependency, even without shuffling, each new branch carries the hidden output of the previous branch, which can reduce the effectiveness of the LSTM's memory capabilities.

The success of ICD can be attributed to its order invariance and the use of labels as input. Feeding the label as input allows the model to construct a latent space from the label and work backwards, effectively teaching the network to derive underlying variables—a strategy that encoder-only models cannot achieve. However, ICD has limitations, including slower inference times and higher data requirements.

Finally, the proposed dataset generation pipeline, including transformations and file conversions, enabled effective training by aligning different conditioning vectors within the same coordinate space.

### 4.4 Visualization results

Figure 3 visualize the absolute error between the label and the model prediction of coronary artery pressure at points on the centerline using one subject from CCTA1000 dataset. Deep green symbolizes little to no error, yellow represents some error, and orange and red represent a lot of error. The better a model performed on a given case, the fewer red spots will be visible. In figure 3, the coronary artery stretches along the left anterior descending (LAD) sub-artery and left circumflex (LCX) sub-artery. Most models struggle near (44, -42, 16) but RD (ours) has the least number of orange spots in that area while keeping a green hue throughout the rest of the artery which indicates little error. The closest competitor, CL, loses accuracy near the LCX. In Figure 3, the proposed diffusion models (a, d) were smoothest only having elevated error near caps.

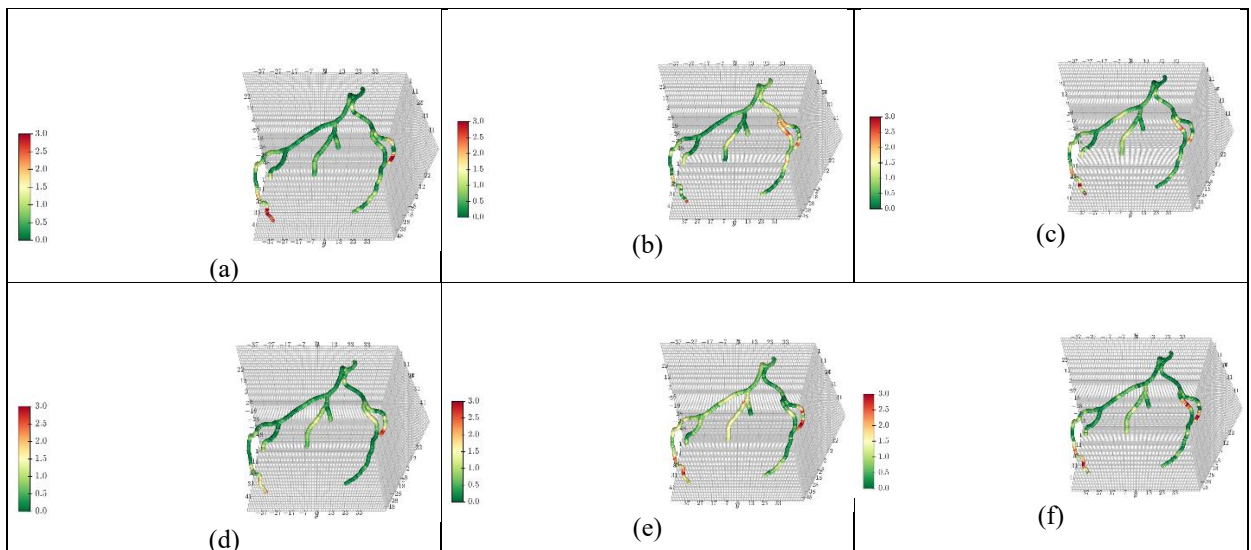

**Figure 3.** Blood Pressure Prediction for one subject in CCTA1000 dataset using (a) RD, (b) RL, (c) RM, (d) CD, (e) CL, (f) CM models. The pseudo bar indicates the difference between the predicted blood pressure and the ground truth of the blood pressure generated during the CFD simulation in mmHg.

Figure 4 visualizes one testing subject in CCTA36 dataset. RL and RM both perform well, being majorly green albeit with a number of red spots at the bifurcations. The attention of the LSTM enables it to maximize the little data available in the FFR dataset.

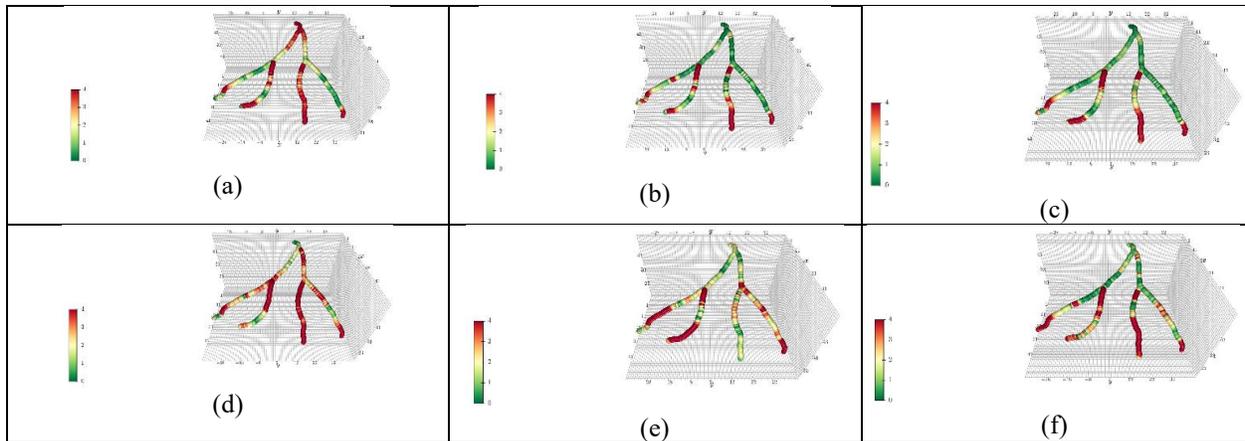

**Figure 4.** Blood Pressure Prediction for one subject in CCTA36 dataset using (a) RD, (b) RL, (c) RM, (d) CD, (e) CL, (f) CM models. The pseudo bar indicates the difference between the predicted blood pressure and the ground truth of the blood pressure generated during the CFD simulation in mmHg.

In Figure 4, the RD and CD models redeem themselves in the CCTA36 dataset by having more consistency along the major arteries, but every model struggles with the branched arteries. This is mostly due to the limited samples in the training dataset. A larger dataset would result in much better performance throughout.

## 5. Conclusion

To summarize, to assist cardiologists in making accurate, non-invasive diagnoses in a timely manner, a novel pipeline and deep learning model were proposed. The deep regressor was a Scalar Diffusion Regressor with Resnet feature extraction. The pipeline was able to document and overcome many challenging obstacles to a way to create a clean and compatible dataset, given only a CT scan and artery segmentation through patch and pressure extraction along a centerline after simulating blood flow and aligning volumes. The model was able to use this data pipeline to predict blood pressure for an entire artery with 64.42% R-squared and leaves room for further model improvement. This study served as the foundation for future research in coronary artery disease detection as well as pushing the envelope further in the current landscape of bioinformatics. The study can be used as an application to non-invasively predict the blood pressure of any person and will hopefully help to diagnose CAD earlier through its implementation.

**Acknowledgement:**

This work is partially supported by American Heart Association under award #25AIREA1377168 (PI. Chen Zhao).

# Supplementary materials

## Section A. CFD simulation steps

### 1. Loading the data

To load the data from a set of raw data, drag and drop the case's CT Scan (.nii image file) and 3D segmentation label (.nii segmentation file) into 3D Slicer's view. The nest steps are to select the segmentation editor, show 3D, then cut the right artery.

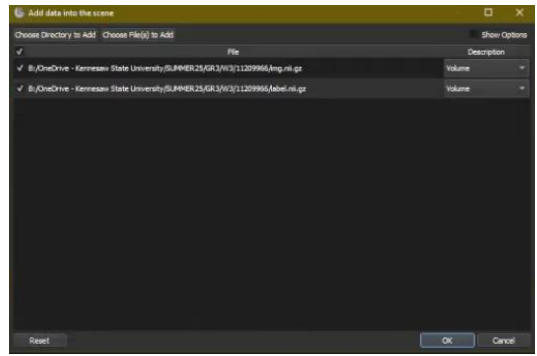

**Figure S1**. Slicer Data Loading

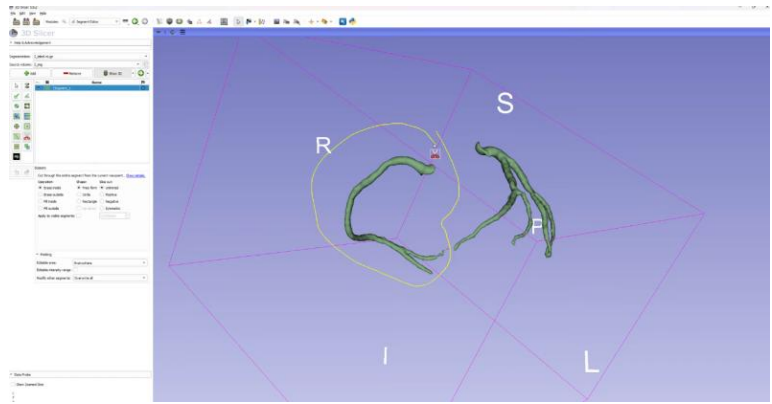

**Figure S2**. Slicer Cutting

### 2. Clipping & Centerline Extraction

After installing the VMTK module in Slicer, in the extract centerline module, set surface to the label file. Set the segment to the existing segment and create a new set of endpoints. Under the tree tab, create a new centerline model then apply. Once applied, adjust the endpoints to ensure that they are only present at the ends of the tree. This will have generated a centerline.

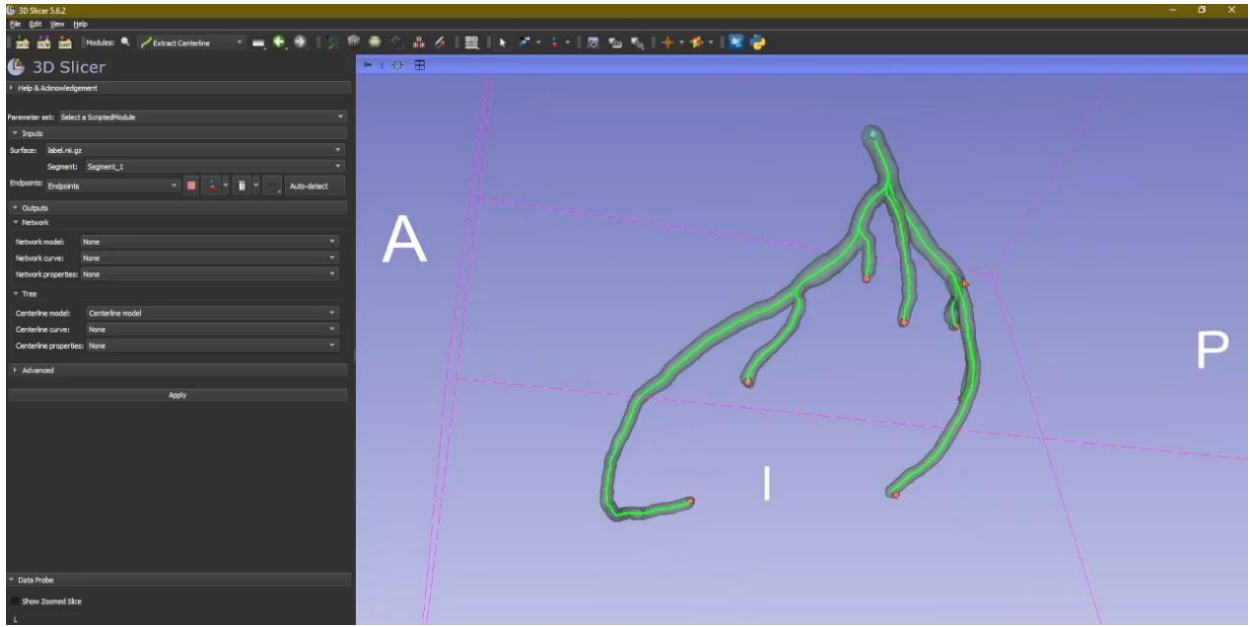

**Figure S3**. VMTK Centerline Extraction

In the clip vessel module, set the "centerlines" to newly generated centerline. For clip points set it to the newly generated endpoints. Under the outputs tab, select a new model. Optionally, the caps output surface can be set to true but false is the standard because the caps will be extracted in the modeling section with more scrutiny. Export the centerline model as a .vtp file and the clipped artery as a .stl file.

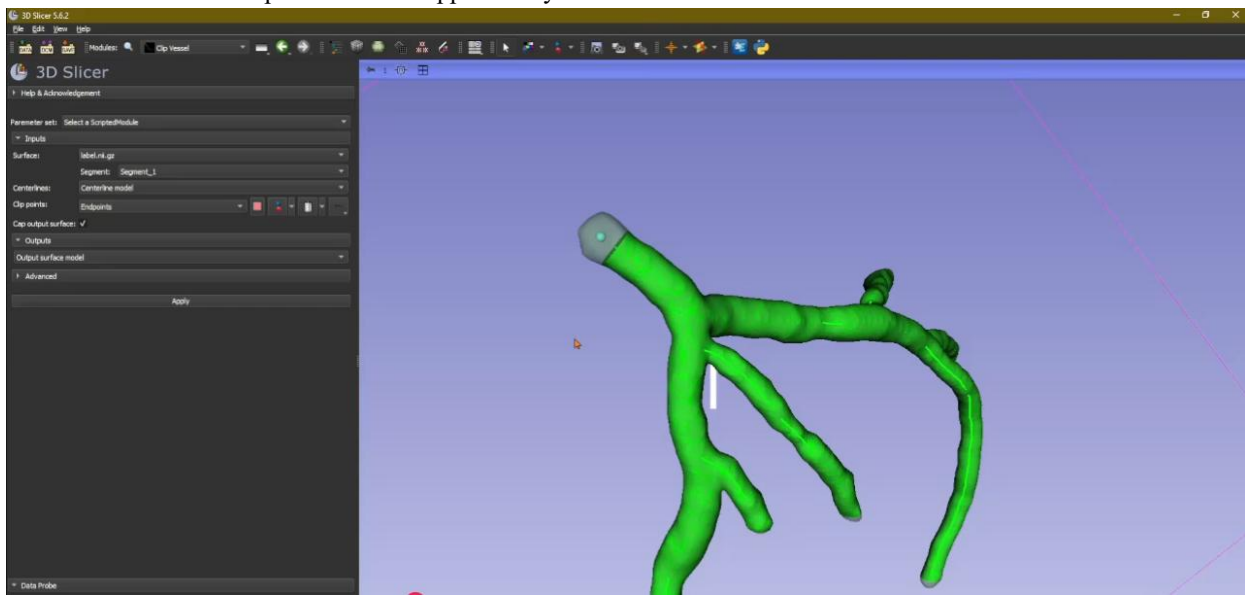

**Figure S4**. VTMK Auto-clipping

## 3. Modeling

In Simvascular, create a new SV project if you do not have one already. Multiple simulations can and should be stored under one SV project. Next, import the clipped artery as a model. You will be prompted to extract the faces, select yes with a separation angle of 50. Reinitialize the global representation to refresh the view. Delete any surfaces that do not correspond to anything. Fill holes with IDs and name the caps according to the artery they belong to like Left Anterior Descending (LAD). Set the type for the surface to wall and set the type for the caps to cap.

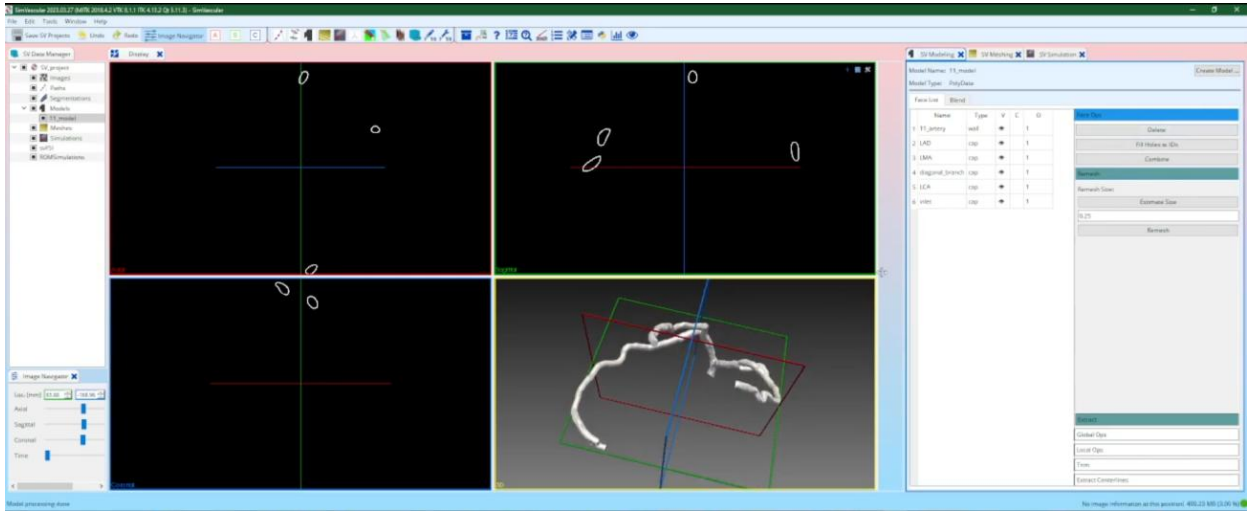

Figure S5. SimVascular Modeling

## 4. Meshing

Make sure to save after every section. Create a new mesh using the newly created model. Estimate the size then run the mesher. If the meshing fails it means that the clip may have a problem and you may need to restart. Sometimes you may just need to re-mesh. To re-mesh revisit the modeling tab and re-estimate size then save. After that, re-run the mesher.

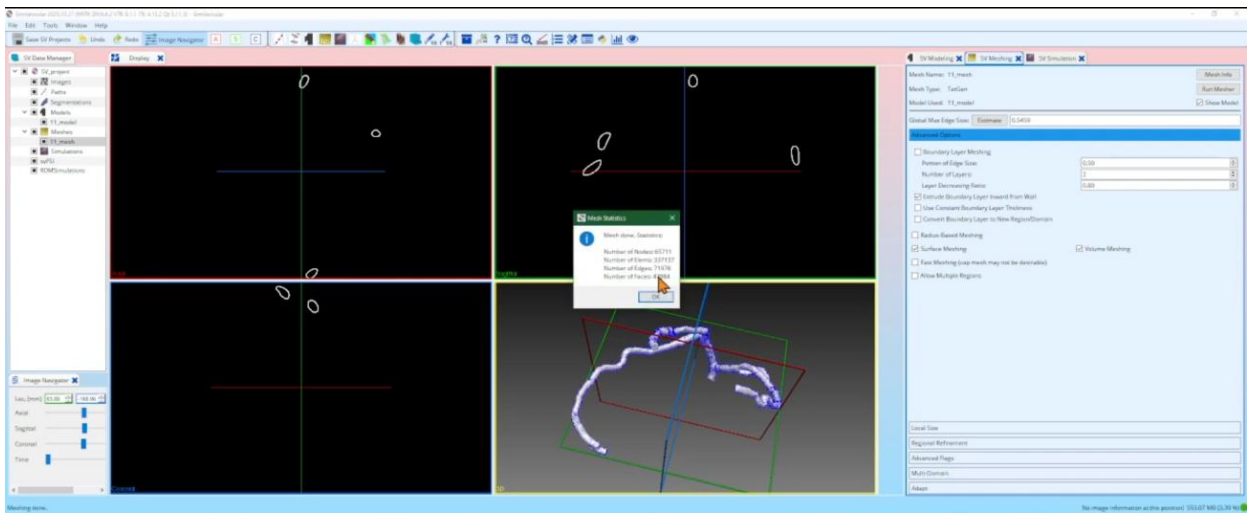

Figure S6. Simvascular Meshing

## 5. Assigning Simulation Parameters

Create a new simulation using the newly generated mesh. For the basic parameters, set the initial pressure to 133300 Pa. The final results will treat this as 100mmHg. Next, for inlet and outlet BCS, set all the caps to a resistance of 1333except the inlet cap. For the inlet cap, create a steadyflow.txt file if you have not already with 0.0 -65 in the first line and 1.0 -65 in the second line. Set the BC type to prescribed values and select steadyflow.txt as the file.

For Wall properties set thickness to 0.2, elastic modulus to 4e6, density to 0.8, and pressure to 133300 Pa.

## 6. Simulation

After configuring the solver parameters, the mesh was reused to generate the simulation input files. When available, the Message Passing Interface (MPI) option was enabled to accelerate computation, and the number of processes was set according to the hardware capability. The blood-flow simulation was then executed to obtain the transient hemodynamic fields. Upon completion, the results were converted into a sequence of outputs spanning 0 to 150 time steps with an interval of 10 steps. These outputs represent the full cardiac cycle and can be visualized using ParaView by importing the generated all_results_xxxxx.vtp files and playing the temporal sequence. For analysis of functional hemodynamics, the average pressure distribution was obtained by loading the average_pressure.vtp file and selecting the pressure_ave_mmhg scalar field for visualization.

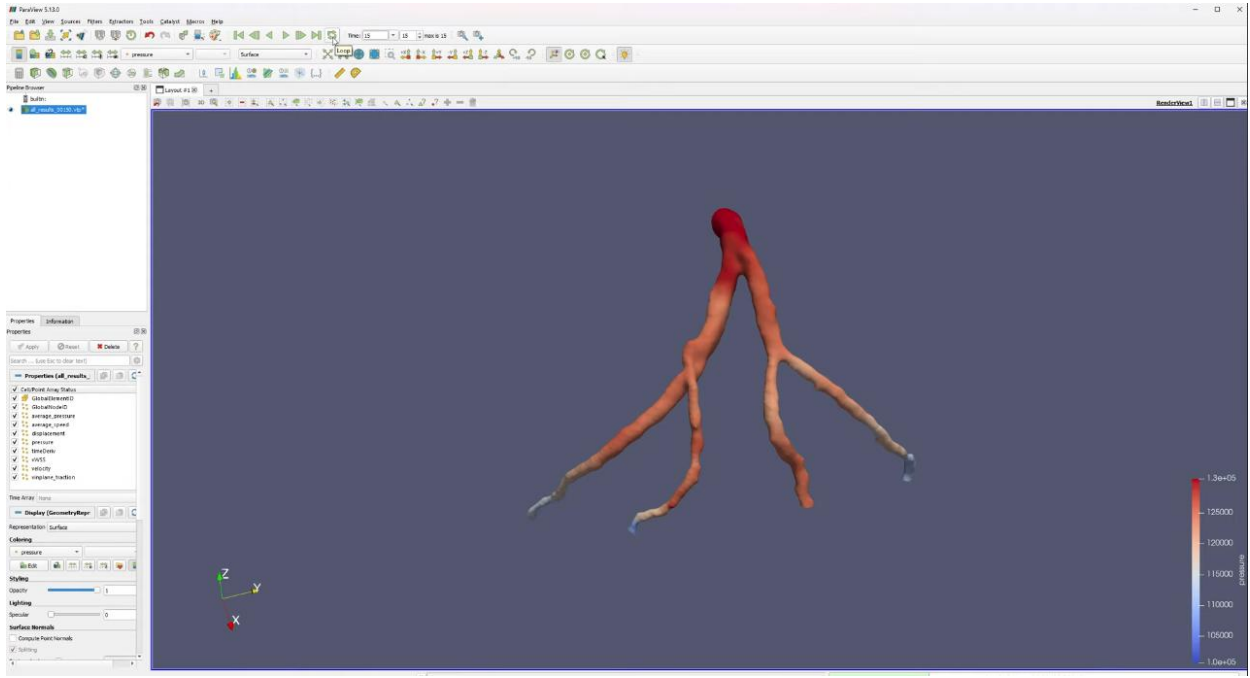

**Figure S7**. Simulation Results in Paraview